\documentclass[conference]{IEEEtran}
\IEEEoverridecommandlockouts

\usepackage{tikz}

\newcommand\copyrighttext{%
  \footnotesize \textcopyright 2025 IEEE. Personal use of this material is permitted.
  Permission from IEEE must be obtained for all other uses, in any current or future
  media, including reprinting/republishing this material for advertising or promotional
  purposes, creating new collective works, for resale or redistribution to servers or
  lists, or reuse of any copyrighted component of this work in other works.}
\newcommand\copyrightnotice{%
\begin{tikzpicture}[remember picture,overlay]
\node[anchor=south,yshift=10pt] at (current page.south) 
  {\fbox{\parbox{\dimexpr\textwidth-\fboxsep-\fboxrule\relax}{\copyrighttext}}};
\end{tikzpicture}%
}

\usepackage{cite}
\usepackage{amsmath,amssymb,amsfonts}
\usepackage{algorithmic}
\usepackage{graphicx}
\usepackage{textcomp}
\usepackage{xcolor}
\usepackage{diagbox}
\def\BibTeX{{\rm B\kern-.05em{\sc i\kern-.025em b}\kern-.08em
    T\kern-.1667em\lower.7ex\hbox{E}\kern-.125emX}}
\begin{document}

\title{Synthetic Dataset Evaluation Based on Generalized Cross Validation
}

\author{\IEEEauthorblockN{1\textsuperscript{st} Zhihang Song}
\IEEEauthorblockA{\textit{Department of Automation} \\
\textit{Tsinghua University}\\
Beijing \\
song-zh22@mails.tsinghua.edu.cn}
\\
\IEEEauthorblockN{4\textsuperscript{th} Lihui Peng}
\IEEEauthorblockA{\textit{Department of Automation} \\
\textit{Tsinghua University}\\
Beijing \\
lihuipeng@tsinghua.edu.cn}

\and
\IEEEauthorblockN{2\textsuperscript{nd} Dingyi Yao}
\IEEEauthorblockA{\textit{Department of Automation} \\
\textit{Tsinghua University}\\
Beijing \\
ydy24@mails.tsinghua.edu.cn}
\\
\IEEEauthorblockN{5\textsuperscript{th} Danya Yao}
\IEEEauthorblockA{\textit{Department of Automation} \\
\textit{Tsinghua University}\\
Beijing \\
 yaody@tsinghua.edu.cn}
 
\and
\IEEEauthorblockN{3\textsuperscript{rd} Ruibo Ming}
\IEEEauthorblockA{\textit{Department of Automation} \\
\textit{Tsinghua University}\\
Beijing \\
mrb22@mails.tsinghua.edu.cn}
\\
\IEEEauthorblockN{6\textsuperscript{th} Yi Zhang}
\IEEEauthorblockA{\textit{Department of Automation} \\
\textit{Tsinghua University}\\
Beijing \\
zhyi@tsinghua.edu.cn}

}

\newcommand{\etal}{\textit{et al.}}

\maketitle
\copyrightnotice
\begin{abstract}
With the rapid advancement of synthetic dataset generation techniques, evaluating the quality of synthetic data has become a critical research focus. Robust evaluation not only drives innovations in data generation methods but also guides researchers in optimizing the utilization of these synthetic resources. However, current evaluation studies for synthetic datasets remain limited, lacking a universally accepted standard framework. To address this, this paper proposes a novel evaluation framework integrating generalized cross-validation experiments and domain transfer learning principles, enabling generalizable and comparable assessments of synthetic dataset quality. The framework involves training task-specific models (e.g., YOLOv5s) on both synthetic datasets and multiple real-world benchmarks (e.g., KITTI, BDD100K), forming a cross-performance matrix. Following normalization, a Generalized Cross-Validation (GCV) Matrix is constructed to quantify domain transferability. The framework introduces two key metrics. One measures the simulation quality by quantifying the similarity between synthetic data and real-world datasets, while another evaluates the transfer quality by assessing the diversity and coverage of synthetic data across various real-world scenarios. Experimental validation on Virtual KITTI demonstrates the effectiveness of our proposed framework and metrics in assessing synthetic data fidelity. This scalable and quantifiable evaluation solution overcomes traditional limitations, providing a principled approach to guide synthetic dataset optimization in artificial intelligence research.

\end{abstract}


\section{Introduction}
The evaluation of synthetic dataset quality plays a crucial role in the development of deep learning\cite{baraheem2023image}. With the significant advancement in hardware computing power, the demand for large-scale datasets in deep learning has become increasingly urgent. As a result, synthetic datasets have emerged as a common approach for generating or augmenting data to support model training and enhance performance\cite{song2023synthetic}. However, the quality of synthetic datasets directly affects the performance and generalization ability of deep learning models. Biases and inaccuracies present in synthetic data may propagate through the training process, potentially leading to erroneous predictions and undesirable outcomes when models are deployed in real-world physical environments. However, research on the evaluation of dataset quality remains limited, and there is currently no widely accepted quantitative method for assessing synthetic datasets. One of the earlier and more commonly used approaches for dataset evaluation is the Analytic Hierarchy Process (AHP)\cite{ahp}. For a given dataset, AHP or its extended version, the Fuzzy Analytic Hierarchy Process (FAHP)\cite{fahp}, allows evaluators to assign manual weights to different evaluation criteria based on expert judgment and to conduct consistency checks to generate an overall score. Nevertheless, this method heavily relies on subjective expert knowledge for indicator selection and weight assignment, which introduces significant bias. As a result, its generalizability and applicability for consistent comparison across diverse datasets are still limited and require further improvement.

In addition, some studies have evaluated the effectiveness of synthetic datasets by examining the training outcomes of models trained on these datasets\cite{park2023study,zhang2023quality,elmquist2024methodology}. To assess the validity and reliability of a dataset, researchers have proposed evaluation criteria such as form quality, content quality, and utility quality\cite{li2023quality}. These metrics aim to capture not only the coverage and representativeness of the data but also the extent to which the dataset can support robust model learning and generalization. Such approaches provide useful insights into the practical utility of synthetic datasets, though standardized evaluation frameworks remain an open research area.

This paper introduces a novel evaluation method for assessing the utility and quality of synthetic datasets by integrating principles from transfer learning and domain adaptation in deep learning. Specifically, we propose an evaluation framework based on general cross-validation experiment, along with two corresponding quantitative metrics. The method is inspired by the inverse process of model transfer learning: by selecting relevant reference datasets and task-specific models, we construct a generalized cross-validation matrix based on the performance of a single model transferred across different target domains. From this matrix, we derive metric weights to produce evaluation scores. This approach enables a more generalizable, quantifiable, and comparable assessment of synthetic dataset quality, addressing the limitations of traditional subjective methods.

\section{Related Works} \label{Relatedworks}
\subsection{Synthetic Datasets for Object Detection}
In the field of object detection, synthetic datasets are playing an increasingly significant role in addressing the challenges of high labeling costs and limited testing scenarios encountered by real-world datasets \cite{song2023synthetic}. 

Virtual KITTI\cite{vkitti} pioneered this approach by reconstructing five real KITTI\cite{kitti} sequences in Unity to produce synthetic driving videos, complete with precise 2D/3D bounding boxes under clear, cloudy, foggy, and heavy‐rain conditions, as well as varying camera viewpoints. Its successor, Virtual KITTI2 \cite{vkitti2}, leverages an updated Unity engine to achieve greater photorealism, incorporates a second‐view camera for binocular RGB output, and preserves the comprehensive multi‐task annotations of the original. Complementing these, VIPER\cite{viper} exploits GTA V’s rendering pipeline to extract 254,064 high‐resolution frames covering 184 km. Each frame is annotated across diverse weather and lighting. ParallelEye\cite{paralleleye} leverages Unity3D to create seven sub‐datasets totaling 40,251 frames under diverse weather and illumination conditions, providing synchronized annotations for detection. At an even larger scale, SHIFT\cite{shift} uses the CARLA simulator to render 2.5 million images across over 4,800 sequences. To address collaborative perception, V2X‐Sim\cite{v2xsim} models vehicle‐to‐infrastructure communication in CARLA\cite{carla} using six cameras with distinct horizontal orientations, yielding 10,000 frames annotated for 3D object detection. Finally, OPV2V\cite{opv2v} offers 11,464 frames containing 232,913 3D vehicle annotations across eight CARLA towns plus a digital Culver City replica, benchmarking up to 16 fusion strategies for cooperative detection.

\subsection{Quality Assessment of Synthetic Image Datasets}

The quality evaluation methods for synthetic image datasets can be categorized into three categories: pixel-level, distribution-level, and task-level. 

Traditional pixel-level metrics, such as Peak Signal-to-Noise Ratio (PSNR)\cite{psnr} and Structural Similarity Index (SSIM)\cite{ssim}, primarily focus on quantifying point-wise similarity between synthetic and reference images in pixel space. 

With the advancement of generative models like generative adversarial networks\cite{gan}, distribution-level evaluation metrics have gradually become mainstream. These metrics leverage pre-trained deep networks to extract high-level semantic features, aiming to measure the alignment between the distributions of synthetic and real images. For instance, the Inception Score (IS)\cite{is} employs an Inception network to compute classification confidence and entropy for generated images, reflecting both image quality and diversity. The Fréchet Inception Distance (FID)\cite{fid} adopts Gaussian fitting of Inception features to calculate the Fréchet distance between the feature distributions of synthetic and real images. Additionally, the Learned Perceptual Image Patch Similarity (LPIPS)\cite{lpips} metric utilizes multi-layer semantic features extracted by pre-trained vision models to construct an image patch similarity measure that aligns with human visual perception. 

For task-level evaluation methods, Park \etal \cite{park2023study} proposed an attention map similarity metric based on ENet-SAD\cite{enetsad} for lane segmentation tasks to quantify Sim2Real gaps in autonomous driving simulations. Zhang  and Eskandarian\cite{zhang2023quality} introduced the Detection Quality Index (DQI), which combines fine-grained saliency maps and object detection results to assess frame-level detection quality in autonomous driving cameras. Elmquist \etal \cite{elmquist2024methodology}  developed a Contextualized Performance Difference (CPD) framework that utilizes semantic alignment and perceptual algorithm performance distributions to measure Sim2Real gaps without paired datasets, applicable to object detection and semantic segmentation tasks.


\section{Methdology}

\begin{figure*}[htbp]
\centerline{\includegraphics[width=0.8\textwidth]{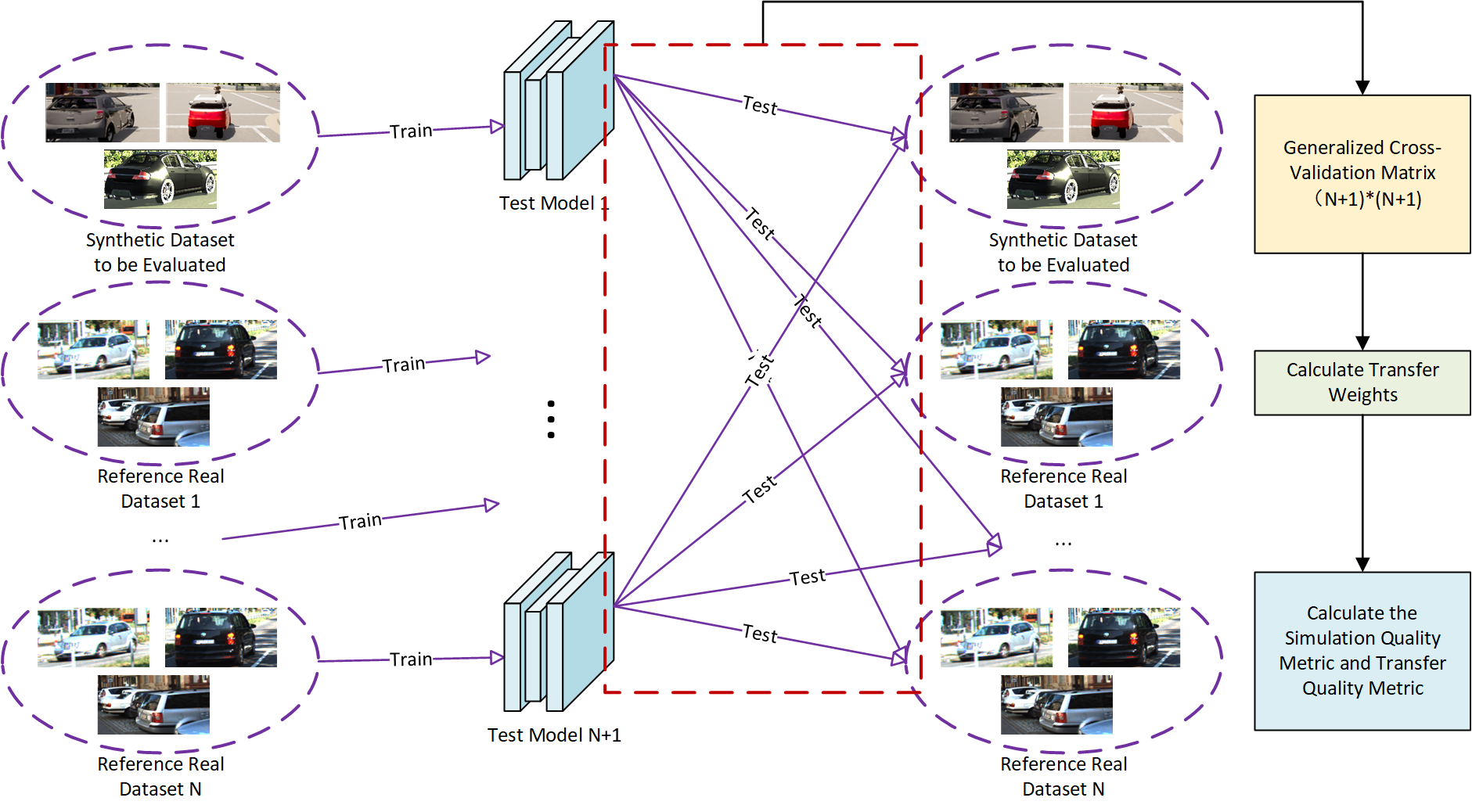}}
\caption{The structure of the generalized cross-validation evaluation framework.}
\label{fig:framework}
\end{figure*}

This section presents a detailed description of the synthetic dataset evaluation method based on the principle of generalized cross-validation. Our method for evaluation synthetic datasets is based on the model transfer test experiments. By utilizing the domain transfer performance of the same model, we can characterize the nature of datasets in turn. In this way, our method can be utilized on various type of datasets by selecting proper task models flexibly and actively according to needs. The structure of our evaluation framework is shown in Figure \ref {fig:framework}.

\subsection{Dataset Preparation}
The evaluation begins with the selection of \( N \) real-world reference datasets \( \{D_i\}_{i=1}^N \), chosen based on their relevance to the target task and alignment with the synthetic dataset \( D_{o} \). These reference datasets are drawn from domains similar to the synthetic data and must either share identical task annotations or exhibit meaningful label overlap. The selection prioritizes datasets that are widely recognized within the target field to ensure the validity and generalizability of the evaluation.  

Following dataset selection, the label spaces of synthetic dataset and the reference datasets are harmonized by identifying their shared categories. Both the synthetic and real-world datasets are then filtered to retain only those samples containing at least one label from shared categories, while non-matching annotations are discarded. This step ensures that subsequent comparisons are performed on semantically consistent subsets of the data. The filtered datasets are partitioned into training and testing sets. To eliminate potential biases arising from dataset size disparities, the training sets are standardized to contain an equal number of samples. This rigorous pre-processing pipeline guarantees that the evaluation metrics derived in later stages reflect the intrinsic quality of the synthetic data rather than artifacts of label mismatch or uneven data volumes. 
\subsection{Evaluation Framework}
The evaluation framework employs a representative deep learning model architecture carefully selected for the target task, such as object detection. This model should either represent state-of-the-art performance in the application domain or constitute a well-established baseline widely adopted in the research community. The key requirements for the model include compatibility with both synthetic and real-world data formats, demonstrated stability in training convergence, and the ability to provide reliable performance metrics across different data distributions.

The evaluation process begins by training the selected model architecture on the preprocessed synthetic dataset $D_o$, using only the shared label categories identified during dataset preparation. The model is trained to convergence using standard optimization procedures, with hyperparameters carefully tuned to prevent overfitting, resulting in the trained model $M_o$. A task-appropriate performance metric $P$ is then selected, such as mean average precision, precision, and recall for detection tasks, to quantitatively assess the model's generalization capability. This metric is first computed on the synthetic dataset's test set, denoted as $P_{oo}$, to establish a baseline performance measure, and subsequently on the test sets of all $N$ reference datasets, recorded as $P_{oi}$ for $i$ = 1,2,...,$N$.

To complete the comprehensive evaluation framework, the same training procedure is repeated for each reference dataset $D_i$, producing $N$ additional trained models $M_i$. Each reference-trained model $M_i$ is then evaluated in two contexts: first on the synthetic dataset's test set, producing $P_{io}$; and then through cross-evaluation on all other reference datasets' test sets, generating $P_{ij}$ for $j$ = 1,2,...,$N$. This rigorous cross-validation strategy generates a total of $(N+1)^2$ performance measurements, forming a complete performance matrix that captures the transferability characteristics between all dataset pairs. The resulting matrix serves as the foundation for subsequent quality metric computation, enabling systematic comparison of the synthetic dataset's fidelity and utility relative to real-world data sources.

\subsection{Performance Matrix Construction}
The evaluation framework systematically organizes the cross-validation results into two key matrices that enable quantitative analysis of dataset transferability and synthetic data quality.

The first is the Cross-Performance Matrix, which captures the absolute performance metrics across all dataset pairs. In this matrix structure, each row corresponds to a model trained on a particular dataset, while each column represents its evaluation performance on another dataset's test set. The matrix's first row and column hold special significance, as they specifically capture how the synthetic dataset performs both in training models and in evaluating models trained on reference datasets. An illustrative example of this matrix based on metric $P$ is shown as Table \ref{cpm}.

\begin{table}[htbp]
    \caption{Cross-Performance Matrix $P$}
    \label{cpm}
    \begin{center}
    \begin{tabular}{ccccccc}
    \hline
        \diagbox{Train set}{Test set} & $O$ & $D_1$ & $D_2$ & $D_3$ & ... & $D_N$ \\ \hline
        $O$ & $P_{oo}$ & $P_{o1}$ & $P_{o2}$ & $P_{o3}$ & ... & $P_{oN}$ \\ 
        $D_1$ & $P_{1o}$ & $P_{11}$ & $P_{12}$ & $P_{13}$ & ... & $P_{1N}$ \\ 
        $D_2$ & $P_{2o}$ & $P_{21}$ & $P_{22}$ & $P_{23}$ & ... & $P_{2N}$ \\ 
        $D_3$ & $P_{3o}$ & $P_{31}$ & $P_{32}$ & $P_{33}$ & ... & $P_{3N}$ \\ 
        ... & ... & ... & ... & ... & ... & ... \\ 
        $D_N$ & $P_{No}$ & $P_{N1}$ & $P_{N2}$ & $P_{N3}$ & ... & $P_{NN}$ \\ \hline
    \end{tabular}
    \end{center}
\end{table}

In addition, to better isolate the intrinsic dataset quality from model-specific performance variations, we introduce the Generalized Cross-Validation (GCV) Matrix through a normalization procedure. This transformation addresses the confounding effects of architectural differences and parameter scales across models by focusing on relative performance changes rather than absolute metric values. 

Specifically, for a model trained on the synthetic dataset and evaluated on a reference dataset, we define the normalized transfer performance as
\begin{equation}
R_{oi} = \frac{P_{oi}}{P_{oo}}\label{eq0}
\end{equation}
where $P_{oi}$ is the performance of the model trained on the synthetic dataset and tested on the $i_{th}$ reference dataset. $P_{oo}$ is the performance of the same model evaluated on the synthetic dataset itself.

This ratio $R_{oi}$ measures the model’s ability to transfer from the synthetic source domain to the $i_{th}$ real-world reference domain, thus reflecting the similarity or substitution between the synthetic and real datasets. Typically, $R_{oi} < 1$, as domain shift often leads to reduced performance on unseen datasets. However, if $R_{oi} > 1$, it suggests that the synthetic dataset has higher coverage, diversity, or quality than the reference dataset and may serve as a superior substitute.

By applying this normalization to all elements of the cross-performance matrix, we construct the Generalized Cross-Validation Matrix, where each element represents the normalized transfer performance of a model trained on one dataset and evaluated on another. 

In this matrix, the first row contains the normalized performance of the model trained on the synthetic dataset across all reference datasets, and is used to derive the final quality evaluation indicators. An illustrative format of the Generalized Cross-Validation Matrix is shown as Table \ref{gcvm}.

\begin{table}[htbp]
    \caption{Generalized Cross-Validation Matrix $G$}
    \label{gcvm}
    \begin{center}
    \begin{tabular}{ccccccc}
    \hline
        \diagbox{Train set}{Test set} & $O$ & $D_1$ & $D_2$ & $D_3$ & ... & $D_N$ \\ \hline
        $O$ & 1 & $R_{o1}$ & $R_{o2}$ & $R_{o3}$ & ... & $R_{oN}$ \\ 
        $D_1$ & $R_{1o}$ & 1 & $R_{12}$ & $R_{13}$ & ... & $R_{1N}$ \\ 
        $D_2$ & $R_{2o}$ & $R_{21}$ & 1 & $R_{23}$ & ... & $R_{2N}$ \\ 
        $D_3$ & $R_{3o}$ & $R_{31}$ & $R_{32}$ & 1 & ... & $R_{3N}$ \\ 
        ... & ... & ... & ... & ... & ... & ... \\ 
        $D_N$ & $R_{No}$ & $R_{N1}$ & $R_{N2}$ & $R_{N3}$ & ... & 1 \\ \hline
    \end{tabular}
    \end{center}
\end{table}

Therefore, the above matrix transforms the transfer performance of the test models into a representation of the domain transferability between the synthetic dataset under evaluation and the reference datasets.

\subsection{Quality Metric Calculation}

To evaluate whether a synthetic dataset is sufficiently realistic or similar to a given real-world dataset, we proposes a simulation quality metric $A_o$ for synthetic datasets, based on the principle of generalized cross-validation. The metric quantifies dataset similarity through a weighted average of relevant transfer elements within the performance matrix, calculated as
\begin{equation}
A_o = \sum_{i=1}^{N} w_i R_{oi}\label{eq1}
\end{equation}
where the weighting coefficient $w_i$ is determined according to the similarity of object detection performance between the synthetic dataset under evaluation and each reference dataset. Given the diversity and domain-specific focus of real-world data across different datasets for the same task, it is not necessary for the synthetic dataset to perform equally well on all real datasets. Rather, if the synthetic dataset demonstrates sufficient similarity with any real dataset within the same domain and task, it can be considered to have high visual realism and annotation fidelity. Therefore, to determine the transfer weight $w_i$, we define
\begin{equation}
R_{io} = \frac{P_{io}}{P_{ii}}\label{eq2}
\end{equation}

\begin{equation}
w_i = \frac{R_{io}}{\sum_j R_{jo}}, \quad j = 1, 2, \dots, N \label{eq3}
\end{equation}
where $R_{io}$ represents the normalized transfer performance of the object detection model, reflecting the similarity of algorithmic performance when transferred from the reference dataset to the synthetic dataset. A higher $R_{io}$ indicates smaller domain shifts and greater similarity. Thus, $w_i$ reflects the relative importance of each reference dataset in evaluating the realism of the synthetic dataset.

In addition to realism, a comprehensive evaluation of a synthetic image dataset should consider its diversity and coverage—specifically, whether the dataset effectively supplements training for various real-world datasets sharing the same label set. To this end, we propose a transfer quality metric $S_o$, based on GCV. This is achieved by computing cross-domain normalized weight coefficients $v_i$ and performing a weighted average of transfer performances under the specified task
\begin{equation}
S_o = \sum_{i=1}^{N} v_i R_{oi}\label{eq4}
\end{equation}

\begin{figure}[htbp]
\centerline{\includegraphics[width=0.5\textwidth]{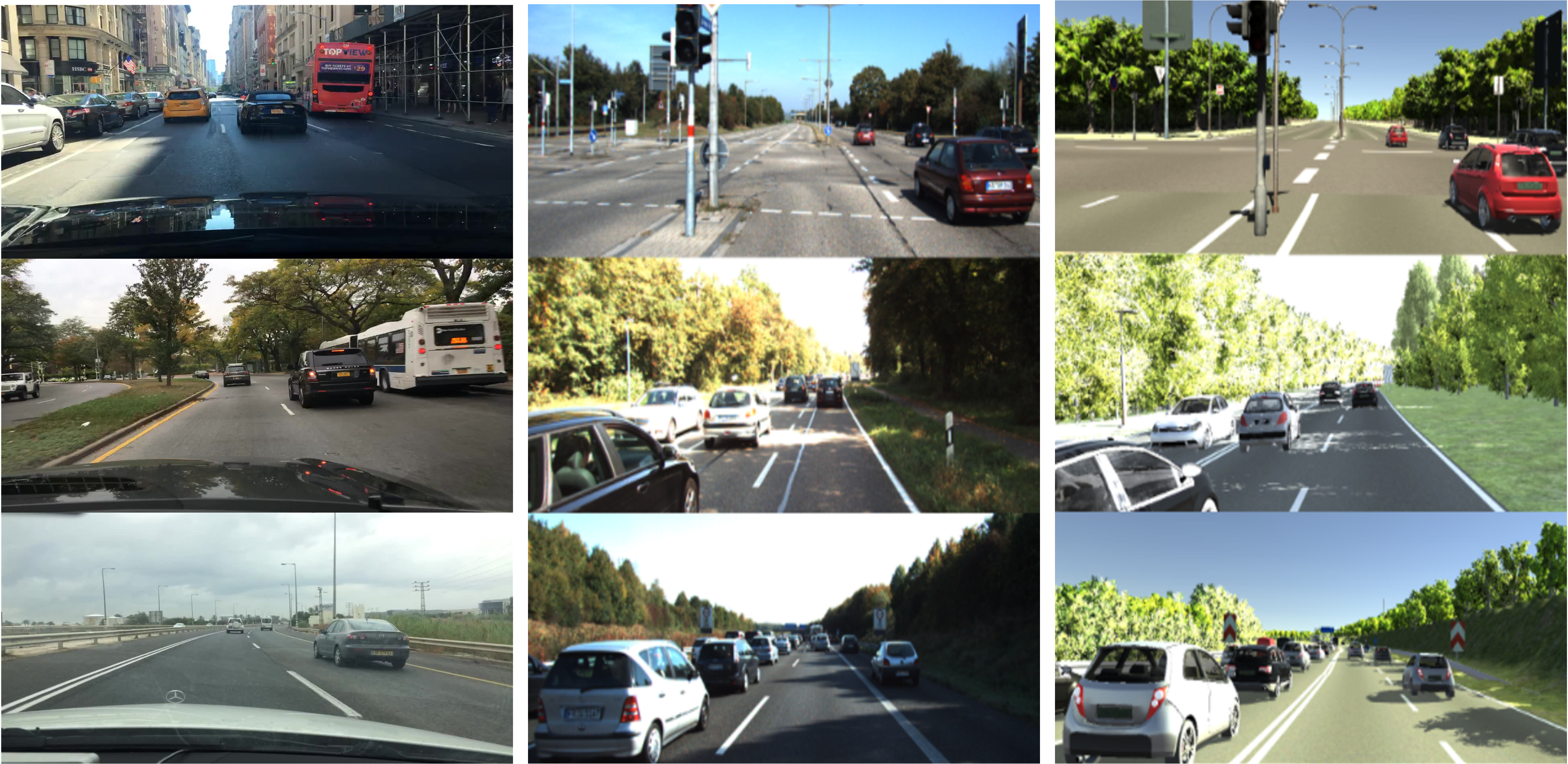}}
\caption{Example of evaluating synthetic dataset and reference real datasets ( left to right: BDD100K\cite{bdd100k}, KITTI\cite{kitti}, and Virtual KITTI\cite{vkitti}). }
\label{fig:datas}
\end{figure}

The weight coefficient $v_i$ is determined based on inter-dataset transfer relationships as follows:
\begin{equation}
R_{ij} = \frac{P_{ij}}{P_{ii}}, \quad i,j = 1,2,\dots,N \label{eq5}
\end{equation}

\begin{equation}
C_{ij} = \frac{R_{ij}}{R_{ij} + R_{ji}}, \quad i,j = 1,2,\dots,N,\ i \neq j \label{eq6}
\end{equation}

\begin{equation}
v_i = \frac{2}{N-1} \sum_{i \ne j} C_{ij} \label{eq7}
\end{equation}
where $R_{ij}$ reflects the transfer performance from dataset $D_i$ to $D_j$. A higher value indicates stronger similarity or coverage of $D_j$ by $D_i$. The coefficient $C_{ij}$ quantifies the directional importance of the transfer relationship between $D_i$ and $D_j$, with larger values implying higher transfer dominance of $D_i$. Due to its symmetric nature, $C_{ij}$ captures the mutual relationship among different reference datasets.

Finally, the transfer weights $v_i$ are normalized sums of $C_{ij}$ for each reference dataset as a source domain, thereby reflecting its centrality in the inter-dataset transfer network. A larger $v_i$ indicates that the synthetic dataset's performance on the corresponding reference dataset better represents its general transferability. For consistency across different evaluations, the final transfer quality metric $S_o$ is also normalized.

\section{Experiments}

Since synthetic dataset Virtual KITTI\cite{vkitti} is constructed based on real-world scenes from the KITTI\cite{kitti}, it provides an intuitive case for evaluating both the quality and transferability of synthetic data. This also enables an assessment of how well the synthetic dataset replicates the characteristics of its real counterpart. Therefore, in our experiments, we select KITTI as a primary reference dataset for evaluating the realism of Virtual KITTI. Additionally, we incorporate BDD100K dataset\cite{bdd100k}, a large-scale open-source real-world dataset for autonomous driving, as a supplementary reference dataset to evaluate generalization and diversity. A visual comparison of the three datasets is shown in Figure \ref{fig:datas}.

In the experiments, we select vehicle as the shared object detection category across all three datasets. The image data and annotations of each dataset are filtered and cleaned to retain only relevant images and labels associated with this category. Based on the size of the filtered target dataset, we set the training set size to 7,400 images and applied the same split ratio to the other datasets.

The task chosen for evaluation is object detection in autonomous driving scenarios, a critical perception task that has seen significant advancements in recent years\cite{ku2019monocular,liu2021autoshape,dong2025novel}. We work adopt the well-established YOLOv5s\cite{yolov5}, a representative model from the YOLO series of object detectors, as the baseline model. The evaluation metric used was the Average Precision at IoU threshold 0.5 (AP50).

Three YOLOv5s models were independently trained on the training sets of the synthetic dataset (Virtual KITTI) and the two real-world datasets (KITTI and BDD100K), respectively. Each model was then evaluated on the test sets of all three datasets to obtain the cross-domain performance matrix as shown in Table \ref{exp1}.

\begin{table}[htbp]
    \caption{Virtual KITTI Cross-Performance Matix}
    \label{exp1}
    \begin{center}
    \begin{tabular}{cccc}
    \hline
        \diagbox{Train set}{Test set} & Virtual KITTI & KITTI & BDD100K\\ \hline
        Virtual KITTI & 0.936 & 0.618 & 0.365\\ 
        KITTI & 0.759 & 0.985 & 0.256\\
        BDD100K & 0.885 & 0.642 & 0.712\\ \hline
    \end{tabular}
    \end{center}
\end{table}

Thus, we can further get the generalized cross-validation matrix as shown in Table \ref{exp2}.

\begin{table}[htbp]
    \caption{Virtual KITTI generalized cross-calidation matrix}
    \label{exp2}
    \begin{center}
    \begin{tabular}{cccc}
    \hline
        \diagbox{Train set}{Test set} & Virtual KITTI & KITTI & BDD100K\\ \hline
        Virtual KITTI & 1 & 0.66 & 0.39\\ 
        KITTI & 0.77 & 1 & 0.26\\
        BDD100K & 1.24 & 0.9 & 1\\ \hline
    \end{tabular}
    \end{center}
\end{table}

From the generalized cross-domain verification matrix, it can be observed that Virtual KITTI and KITTI datasets demonstrate relatively good bidirectional cross-domain performance, significantly outperforming the performance observed when transferring to the other real-world dataset, BDD100K. This result aligns with the design of the Virtual KITTI synthetic dataset, which was created to simulate the characteristics of KITTI, and shows notable differences compared to the scenarios captured in other real-world datasets.

Based on the generalized cross-domain matrix above, we calculate the weighted transfer scores. According to  (\ref{eq1}), the simulation quality indicator $A_o$ for the synthetic object detection dataset is derived as

\begin{equation}
A_o = \frac{0.66 \times 0.77}{1.24 + 0.77} + \frac{0.39 \times 1.24}{1.24 + 0.77} = 0.49\label{eq10}
\end{equation}

We then calculate the multi-cross transfer normalized weighting coefficients and obtain the transfer quality indicator $S_o$ for the synthetic object detection dataset using the generalized cross-verification approach according to (\ref{eq4}).

\begin{equation}
S_o =  \frac{ 0.26}{0.26 + 0.90}\times 0.66 +  \frac{ 0.90}{0.26 + 0.90} \times 0.39= 0.45\label{eq12}
\end{equation}

To summarize, under the vehicle detection task benchmarked against the KITTI and BDD100K datasets, the synthetic dataset Virtual KITTI achieves a simulation quality score $A_o = 0.49$ and a transfer quality score $S_o = 0.45$ according to the generalized cross-domain verification method.

\section{Conclusion}
This study presents a generalized cross-validation based evaluation framework for synthetic datasets, addressing the critical challenge of quantifying domain transferability and generalizability. By integrating task-specific model training across synthetic and real-world datasets, the proposed approach constructs a standardized evaluation matrix that overcomes the subjectivity and fragmentation of traditional methods. The defined metrics \(A_o\) and \(S_o\) enable systematic assessment of synthetic data fidelity and its applicability across diverse real-world scenarios, as validated by the Virtual KITTI experiment where the framework achieved \(A_o = 0.49\) and \(S_o = 0.45\) in vehicle detection tasks. The framework’s key contributions lie in its quantifiable and comparable evaluation paradigm. This not only advances the theoretical foundation of synthetic data evaluation but also offers tangible guidance for researchers—enabling data-driven optimization of synthetic generation pipelines and facilitating informed decisions on dataset selection.

The evaluation framework proposed in this paper relies on the downstream task of object detection and assumes overlapping label spaces between synthetic and real datasets. However, some synthetic datasets either lack corresponding annotations or contain non-overlapping label definitions, which limits the applicability of the framework. Future work may explore evaluation methods that do not depend on downstream task labels, thereby enhancing universality and scalability. Future work will also include validating the framework across a broader range of synthetic and real-world datasets to further demonstrate its robustness and general applicability.



\bibliographystyle{IEEEtran}  
\bibliography{conference_101719}

\end{document}